\documentclass[twocolumn]{article}

\usepackage{graphicx}
\usepackage{tikz}
\usetikzlibrary{arrows, positioning, arrows.meta, fit, backgrounds, calc}
\usepackage{booktabs}
\usepackage{amssymb}
\usepackage{amsmath}
\usepackage[breaklinks=true]{hyperref}
\usepackage{xurl}
\usepackage[numbers]{natbib}
\usepackage{microtype}
\usepackage{geometry}
\usepackage{xcolor}
\usepackage{xspace}
\usepackage{stfloats}
\usepackage{float}
\usepackage{caption}
\usepackage{enumitem}

\newcommand{\goodmanGamma}{0.084}

\newcommand{\topOneAcc}{34.6\%}

\newcommand{\sensNPackages}{1695}
\newcommand{\sensNVariants}{6978}
\newcommand{\sensGamma}{0.054}
\newcommand{\sensGammaCILow}{0.025}
\newcommand{\sensGammaCIHigh}{0.083}

\newcommand{\baseGamma}{0.361}

\newcommand{\baseTopOne}{49.2\%}

\newcommand{\hThreeN}{1678}
\newcommand{\hThreeTopOne}{0.059}
\newcommand{\hThreeTopOneCILow}{0.011}
\newcommand{\hThreeTopOneCIHigh}{0.107}

\newcommand{\nBaseRightPersonaWrong}{126}

\newcommand{\stabGammaSD}{0.025}
\newcommand{\stabTopOneSD}{0.008}

\newcommand{\xfN}{399}
\newcommand{\xfPairedN}{395}

\newcommand{\newSelectedN}{190}
\newcommand{\legacySelectedN}{399}
\newcommand{\correctedPanelN}{190}
\newcommand{\correctedBaseN}{189}
\newcommand{\correctedPairedN}{189}
\newcommand{\correctedPanelTau}{0.049}
\newcommand{\correctedBaseTau}{0.325}
\newcommand{\redditPanelTop}{40.3\%}
\newcommand{\redditBaseTop}{30.2\%}
\newcommand{\redditPanelLift}{0.452}
\newcommand{\redditBaseLift}{0.333}

\newcommand{\newGeminiN}{188}
\newcommand{\newGeminiExcludedN}{2}
\newcommand{\newGeminiPanelRankUndefinedN}{3}
\newcommand{\newGeminiBaselineRankUndefinedN}{51}
\newcommand{\newGeminiPanelTau}{0.082}

\newcommand{\newGeminiPanelTopOne}{39.1\%}

\newcommand{\newGeminiBaselineTau}{0.336}

\newcommand{\newGeminiBaselineTopOne}{54.8\%}

\newcommand{\newGeminiGapTau}{+0.222}
\newcommand{\newGeminiGapTauLow}{+0.061}
\newcommand{\newGeminiGapTauHigh}{+0.385}
\newcommand{\newGeminiPairedTauN}{136}
\newcommand{\newGeminiGapTopOne}{+15.7}
\newcommand{\newGeminiGapTopOneLow}{+7.1}
\newcommand{\newGeminiGapTopOneHigh}{+24.0}

\newcommand{\newGptN}{188}
\newcommand{\newGptExcludedN}{2}

\newcommand{\newGptPanelTau}{0.088}

\newcommand{\newGptPanelTopOne}{42.0\%}

\newcommand{\newGptBaselineTau}{0.345}

\newcommand{\newGptBaselineTopOne}{61.7\%}

\newcommand{\newGptGapTau}{+0.257}
\newcommand{\newGptGapTauLow}{+0.119}
\newcommand{\newGptGapTauHigh}{+0.391}
\newcommand{\newGptPairedTauN}{188}
\newcommand{\newGptGapTopOne}{+19.7}
\newcommand{\newGptGapTopOneLow}{+10.6}
\newcommand{\newGptGapTopOneHigh}{+28.7}

\newcommand{\riskCorpusN}{318}
\newcommand{\riskLegacyN}{83}
\newcommand{\riskSelectedN}{42}
\newcommand{\windowGeminiN}{148}
\newcommand{\windowGeminiExcludedN}{40}
\newcommand{\windowGeminiGapTau}{+0.157}
\newcommand{\windowGeminiGapTauLow}{-0.030}
\newcommand{\windowGeminiGapTauHigh}{+0.346}
\newcommand{\windowGeminiPairedTauN}{109}
\newcommand{\windowGeminiGapTopOne}{+11.4}
\newcommand{\windowGeminiGapTopOneLow}{+1.7}
\newcommand{\windowGeminiGapTopOneHigh}{+21.0}

\newcommand{\windowGptN}{147}
\newcommand{\windowGptExcludedN}{41}
\newcommand{\windowGptGapTau}{+0.272}
\newcommand{\windowGptGapTauLow}{+0.118}
\newcommand{\windowGptGapTauHigh}{+0.425}
\newcommand{\windowGptPairedTauN}{147}
\newcommand{\windowGptGapTopOne}{+22.4}
\newcommand{\windowGptGapTopOneLow}{+12.2}
\newcommand{\windowGptGapTopOneHigh}{+32.7}

\title{Do Synthetic Personas Predict Real Audience Response?\\
A Headline-Ranking Study with a Generic-Reader Baseline}

\author{Alexandre Cristov\~ao Maiorano\\
\texttt{alexandre@lumytics.com}}
\date{}

\begin{document}

\twocolumn[
  \begin{@twocolumnfalse}
    \maketitle
    \begin{abstract}
Do demographic personas improve an LLM's prediction of audience engagement? We compare a fixed
ten-persona panel with a generic-reader baseline on Upworthy headlines, with supplementary
news and Reddit comparisons. The primary experiment collects Gemini 3.1 Flash-Lite and GPT-4.1
observations on the same \newSelectedN{}-package exploratory cohort, selected with simultaneous
winner screening. It allocates thirty scores per arm and variant within model, retains exact raw
responses, and reports metric-specific valid and paired counts. Complete-package coverage is
\newGeminiN{} for Gemini and \newGptN{} for GPT-4.1. The baseline-minus-panel Kendall $\tau_b$ gap is
\newGeminiGapTau{} (CI [\newGeminiGapTauLow{}, \newGeminiGapTauHigh{}]) for Gemini and
\newGptGapTau{} (CI [\newGptGapTauLow{}, \newGptGapTauHigh{}]) for GPT-4.1, on respectively
\newGeminiPairedTauN{} and \newGptPairedTauN{} valid rank pairs. Expected top-1 is
\newGeminiPanelTopOne{} versus \newGeminiBaselineTopOne{} for Gemini and \newGptPanelTopOne{}
versus \newGptBaselineTopOne{} for GPT-4.1 (panel versus baseline). These intervals condition on
the selected cohort and recorded generations and do not adjust for dependence between related experiments. The Upworthy archive documents a randomization-risk
period. Excluding that period in a post-hoc sensitivity retains positive expected top-1 and lift
gaps in both configurations, while Gemini tau-b and Spearman intervals include zero. Supplementary
Reddit top-1 and captured lift favor the panel; these aggregate comparisons use unreconciled
collection protocols and do not isolate dataset effects. A raw-replay artifact supports offline reproduction of the primary experiment.
The results concern this panel, cohort, and protocol; they are not a held-out evaluation or
evidence of universal persona harm. Provider request protocols differ across models, and no
causal mechanism is isolated.
\end{abstract}

    \vspace{2mm}
  \end{@twocolumnfalse}
]

\section{Introduction}
LLM-based audience simulation offers a way to compare messages before live testing.
Demographic conditioning can approximate aspects of survey samples~\citep{argyleSiliconSampling},
but a copy-ranking system must predict measured engagement. We study this narrower question with
Upworthy headline experiments~\citep{matiasUpworthy}.

We compare a fixed ten-persona panel with a generic-reader baseline using a matched-budget
experiment on a simultaneously screened Upworthy cohort. Both model configurations allocate the
same number of scores to each arm and variant, preserve raw responses, and report distinct rank
and decision metrics with explicit denominators. The primary analysis uses uniform expected
credit for tied decisions and per-metric paired contrasts; a post-hoc period-exclusion sensitivity
examines the archive's documented randomization risk. The cohort contains previously examined
outcomes, so this is an exploratory evaluation rather than a held-out test.

Supplementary evidence includes aggregate-report comparisons and a GPT-4.1 scalar-checkpoint
analysis. These use different collection budgets, winner screens, and tie rules and are reported
separately from the primary comparison. Their evidential scope also differs: reexporting an
aggregate does not reconstruct its raw observations, and replaying stored scalars cannot validate
absent response texts. Raw-backed reproduction is available for the primary comparison; it does
not validate the supplementary generations or the other datasets.

We ask (RQ1) how this panel and generic baseline compare within tested packages, (RQ2) how
agreement varies with the observed outcome signal, and (RQ3) what the dataset, aggregation,
seed, and model comparisons establish. This evaluates a specified harness rather than all
persona methods or audience representativeness.

\section{Related Work}
Profile-conditioned models can approximate some aggregate survey distributions
\citep{argyleSiliconSampling}; broader computational social science applications raise validity
questions~\citep{ziems2024css}. Interview-grounded agents reproduce General Social Survey responses
at approximately 85\% of participants' own test--retest consistency in the original version of
\citet{park2024agents}. That result uses rich individual grounding and a different outcome from clicks.

Other work documents option-order sensitivity~\citep{dominguezolmedo2024questioning}, limited effects
of persona attributes on subjective annotation, alongside modest significant gains in some
settings~\citep{hu2024persona}, persona-related degradation
on objective tasks~\citep{zheng2024helpful}, caricature~\citep{cheng2023compost}, and biased outcomes
from generated personas~\citep{li2025personacatch}. Reasoning biases in persona-assigned models are also documented~\citep{gupta2024bias}. These motivate a baseline, but do not explain
this study's gap.

Upworthy~\citep{matiasUpworthy} supports engagement analyses including negativity effects
\citep{robertson2023negativity}. LOLA~\citep{ye2024lola} studies LLM-assisted content experiments;
differences in selection, models, and evaluation prevent a direct numerical comparison here.
MIND~\citep{wu2020mind} and Reddit reposts~\citep{lakkaraju2013name} are additional observational
settings, not interchangeable headline experiments. Construct validity~\citep{cronbach1955construct}
distinguishes intent from behavior. Gamma~\citep{goodman1954} must also be distinguished from
Kendall correlation and its treatment of ties~\citep{kendall1938,kendall1945ties}.

\section{Sim-to-Real Validity Framework}
A package contains variants with observed outcome rates and predicted scores
(Figure~\ref{fig:protocol}). Upworthy was designed for randomized traffic, but the archive authors document randomization
problems during part of 2013--2014~\citep{matias2024correction}. We report observed rate rankings;
they are not automatically causal headline effects. Grouping distinct headlines can also combine
image variants. MIND ranks different articles within a subcategory by clicks divided by logged
article appearances, pooling user-level recommendation logs. Reddit groups reposts by image and
subreddit, pools votes for identical titles, and uses upvotes divided by total upvotes plus
downvotes. This is a vote-approval ratio, not CTR; vote totals are not exposure counts. Timing,
recommendation, and exposure remain uncontrolled in these observational comparisons.
% Authored schematic; no empirical numbers or external data.
\begin{figure*}[t]
\centering
\begin{tikzpicture}[
  font=\small,
  box/.style={draw, rounded corners, align=center, inner sep=5pt, text width=2.7cm, minimum height=12mm},
  >=Stealth, node distance=7mm and 8mm
]
\node[box,fill=green!8] (data) {Observed variants\\and outcomes};
\node[box,fill=blue!6,right=of data] (screen) {Winner\\screening};
\node[box,fill=blue!6,above right=2mm and 8mm of screen] (panel) {Persona panel\\mean score};
\node[box,fill=blue!6,below right=2mm and 8mm of screen] (base) {Generic reader\\mean score};
\coordinate (mid) at ($(panel.east)!0.5!(base.east)$);
\node[box,fill=gray!8,right=8mm of mid,anchor=west] (metrics) {Within-package\\ranking metrics\\and bootstrap CIs};
\draw[->] (data) -- (screen);
\draw[->] (screen.east) -- (panel.west);
\draw[->] (screen.east) -- (base.west);
\draw[->] (panel.east) -- (metrics.north west);
\draw[->] (base.east) -- (metrics.south west);
\end{tikzpicture}
\caption{Comparison protocol. Outcomes determine selection; the two scoring arms are compared
within selected packages. Supplementary aggregates use gamma and the top-versus-runner-up screen. The primary experiment
and GPT checkpoint analysis additionally report tau-b and simultaneous selection. Selection does not establish
that excluded packages are unpredictable.}
\label{fig:protocol}
\end{figure*}

Within-package ranking does not establish absolute CTR calibration or a causal copy effect
in observational data.

\paragraph{Metric identity.}
For concordant and discordant pair counts $C,D$, Goodman--Kruskal gamma is
\[
\gamma=(C-D)/(C+D),
\]
discarding pairs tied in either variable~\citep{goodman1954}. It is undefined when $C+D=0$ and
differs from Kendall $\tau_b$, which accounts for ties:
\[
\tau_b=\frac{C-D}{\sqrt{(C+D+T_x)(C+D+T_y)}},
\]
where $T_x,T_y$ count pairs tied only in the respective variable~\citep{kendall1945ties}.
Spearman uses average ranks~\citep{spearman1904}. Constant rank inputs yield undefined values.

The primary experiment uses uniform expected top-1 and bottom-1 credit:
$|S_{\mathrm{pred}}\cap S_{\mathrm{real}}|/|S_{\mathrm{pred}}|$ for the corresponding extreme
sets. Captured lift averages observed rates across tied predicted maxima, then scales distance
from the observed minimum by the observed range. Constant outcomes make lift undefined.
Supplementary aggregate-report top-1 results use the first maximum and therefore depend on variant order.

\section{Method Under Test}
The study harness scores each headline with either a demographic persona or a generic reader.
It operates independently of any commercial product. The fixed ten-persona JSON takes inspiration
from Upworthy-era U.S.\ social-media demographics~\citep{holcomb2013newsuse}. These descriptors
do not specify a deterministic behavioral-profile generator or establish representativeness.

\paragraph{Elicitation and aggregation.}
The prompt requests JSON with a \texttt{click\_intent} number in $[0,1]$
(Appendix~\ref{sec:appendix}). Scores are averaged across recorded observations:
\[
\widehat{s}(v)=\frac{1}{|O_v|}\sum_{(i,d)\in O_v}y_{i,d}(v).
\]
With complete draws this equals the mean over personas and draws. With missing draws it can
differ from the supplementary aggregation mean, which weights draw-collapsed personas equally. The supplementary aggregation comparison evaluates these alternative rules.

\paragraph{Sampling and parsing.}
The primary experiment uses temperature $0.8$, three draws per persona and thirty generic-reader
draws per variant, with request seeds recorded for both providers. The supplementary Gemini
comparison uses the same nominal draw allocation. The GPT-4.1 scalar-checkpoint analysis instead
uses three draws per persona versus ten baseline draws; its requests sent no generation seed. Seeds alone do not guarantee identical
provider responses over time.

Supplementary scalars were produced by a permissive parser that could extract a first number,
clamp values, and accept booleans; event frequencies are unknown. The matched-budget collector
accepts only a finite, non-boolean JSON number in range under the exact requested key. Invalid
parses are recorded as terminal missing observations. Configuration hashes and separate protocol
namespaces prevent mixing primary checkpoints with supplementary scores.

\section{Experimental Setup}
\paragraph{Corpus.}
Upworthy rows are grouped by experiment and distinct headline, aggregating identical text.
Packages require at least two headlines and $3{,}000$ impressions per headline; sampling uncertainty
matters in online experiments~\citep{kohavi2009controlled}. The locked exploratory corpus has
\sensNVariants{} variants in \sensNPackages{} packages. In these locked inputs, headline linebreaks
were removed before grouping. One retained variant has such a linebreak in the source CSV; its
package is outside both selected cohorts. Exact locked strings define the saved experiments,
and the original manifests remain unchanged.

\paragraph{Archive allocation risk and sensitivity.}
The archive authors' 2024 report identifies a randomization-risk window from 25 June 2013
through 10 January 2014 inclusive~\citep{matias2024correction}. The preserved CSV contains creation dates but neither the documented problem flag nor a
risk indicator. Calendar-date overlap identifies \riskCorpusN{} corpus packages, \riskLegacyN{}
supplementary-screen packages and \riskSelectedN{} packages in the primary selected cohort.
These counts mark window membership, not individually proven defects. Bonferroni screening
cannot repair allocation problems.

A post-hoc offline sensitivity, specified after the primary collection, excludes an entire experiment if any row for that experiment in the preserved CSV has a
creation date that falls
in that inclusive window, using the original date string without inventing a timezone. Each
model otherwise retains its complete primary-package policy and the same recorded generations,
metrics and bootstrap procedure. Primary reports are preserved unchanged. Dates outside the
window are not a certification of valid randomization; this sensitivity is not confirmatory
or an independent replication.

\paragraph{Simultaneous winner screen.}
The primary experiment requires a unique observed maximum and compares it with every opponent using
two-sided pooled $z$ tests. For $k$ variants the comparison family contains $m=k(k-1)/2$ unordered
pairs. All winner--opponent comparisons must pass $\alpha/m$, with $\alpha=0.05$.
The screening statistic is $\min(1,m\max_{j\ne w}p_{wj})$. Bonferroni's inequality motivates the
adjustment~\citep{nistBonferroni}. The nominal bound is approximate and per package, conditional on
independent binomial observations, fixed sampling horizons, and adequate normal approximation. It is not an across-package
false-discovery guarantee or proof of a population winner. Repeated exposures and Reddit voting
further limit these assumptions. In the locked corpus \newSelectedN{} packages pass, all within
the supplementary selected set described below. A stricter screen does not demonstrate that removed packages had
false winners. The archive documents editorial stopping based on accumulating results~\citep{matiasUpworthy}.
Fixed-horizon $p$-values need not remain calibrated under that stopping policy; Bonferroni does
not repair invalid input $p$-values. The screen defines a reproducible cohort, without a verified
five-percent winner-error guarantee for these archived tests.

\paragraph{Supplementary winner screen.}
The supplementary aggregate comparisons use the observed best variant and runner-up,
comparing them using a one-sided pooled two-proportion $z$ test at $p<0.05$, selecting \legacySelectedN{} packages.
The direction was chosen from the data: even a two-arm null comparison can pass in either
direction, so the nominal one-sided threshold does not imply a five-percent false-positive rate.
Comparing only the runner-up need not separate a winner from every opponent. Outcome selection
changes the evaluation population. Packages failing a screen are not thereby unpredictable.

\paragraph{Matched-budget collection.}
We froze the protocol and its source before collection. Both Gemini 3.1 Flash-Lite and Azure GPT-4.1 score
every variant of the same \newSelectedN{}-package simultaneously screened exploratory cohort. This cohort
was selected from previously examined outcomes and is not a held-out population. The
matched-budget protocol is evaluated on this Upworthy cohort only; MIND, Reddit, other Upworthy
splits, model tiers, seeds and aggregation rules are covered by the supplementary analyses,
which do not establish prompt robustness.

Each variant is allocated three observations for each of ten fixed personas and thirty generic-reader
observations. Package and variant order is shuffled with a recorded seed; baseline/persona
requests are interleaved with randomized order within pairs. Temperature is $0.8$, JSON output
is requested, and output is limited to 64 tokens per choice. Gemini uses single-choice requests
with MINIMAL thinking. Azure uses three indexed choices per request in both arms: ten requests
per arm and variant. We match the number of retained scores within each model, not an identical
transport or inference protocol across providers. Token use and monetary cost are not equalized.
Choices in one request are not assumed
independent. Request seeds, exact prompts, responses, returned model identities, usage, and
failures are preserved. Provider aliases and best-effort seeds do not establish future replay
of generations.

The primary analysis retains a package only when every planned observation is valid in both
arms for all its variants. This also makes equal-persona and observation-weighted panel means
identical. Failures are never replaced with additional semantic draws; bounded retries address
transient transport failures only. Available-observation arm summaries are reported separately as sensitivity
analyses, with their own coverage, in the \texttt{available\_observation\_sensitivity} field of
the committed Gemini and Azure reports in \texttt{recollection/2026-10-31/}. The primary policy is stricter than the scalar-checkpoint analysis
requirement of at least one valid score per variant. Operational compatibility probes are retained
separately and excluded. An Azure prompt refusal initially triggered a global HTTP400 stop. Before comparative analysis,
a documented technical amendment classified only the confirmed content-filter refusal type as
terminal missing while allowing unrelated planned requests to continue. The blocked request is
not retried or rewritten; filtering remains active. Original journal bytes, the pause, source
versions and the amendment marker are preserved. Other permanent failures and budget stops remain
blocking. This is a response to an observed technical incident, not a preregistered amendment.
The fixed cohort closes before inspecting comparative performance;
there is no outcome-dependent stopping or choice of model after observing results.

\paragraph{Estimation and denominators.}
Primary package metrics are averaged without size weighting. The displayed means describe the recorded
packages. Bootstrap intervals summarize between-package resampling variability conditional on
the selection rule and recorded scores; interpreting them for a population of similarly selected
experiments additionally assumes that these packages are exchangeable independent units. Raw-backed reports include
$n_{\mathrm{total}},n_{\mathrm{valid}},n_{\mathrm{undefined}}$ and metric-specific package IDs.
Each paired gap uses packages valid in both arms for that metric. Empty estimates are undefined,
not zero. Primary and period-exclusion percentile bootstrap CIs use $10{,}000$ package resamples, seed $42$
\citep{efron1979bootstrap}. They condition on selection and recorded scores and omit generation,
selection, and outcome-measurement uncertainty. These are pointwise estimation intervals, not
simultaneous guarantees across metrics or models. Resampling treats experiment IDs as independent
units. The archive also records follow-up tests with shared content~\citep{matiasUpworthy};
related IDs may therefore be dependent. The intervals do not adjust for such dependence, and
their coverage as population confidence intervals is not established. Supplementary aggregate coefficient-specific valid counts
are not fully recorded; coverage totals are not those denominators.

\paragraph{Planning and robustness scope.}
We report coefficient estimates and intervals without assigning standardized effect-size
categories~\citep{cohen1988power}. The analysis does not establish prospective power for
averages of within-package coefficients or equivalence. Time-stamped preregistration of all thresholds is not established.

Supplementary reports cover three Upworthy splits, MIND~\citep{wu2020mind},
Reddit~\citep{lakkaraju2013name}, three Gemini models, seed runs, and aggregation. Prompt robustness is not evaluated by these artifacts. Reddit's manifest records many
more selected packages than the evaluated subset; its subset-selection procedure is undocumented.

\section{Results}
\paragraph{Primary matched-budget comparison (RQ1).}
Tables~\ref{tab:new_gemini} and \ref{tab:new_azure} give the primary matched-budget experiment.
Supplementary comparisons below retain their own protocols and evidence boundaries. The complete-package policy retains
\newGeminiN{} Gemini packages and \newGptN{} GPT-4.1 packages; exclusions are respectively
\newGeminiExcludedN{} and \newGptExcludedN{}. Reports retain every planned observation and
failure, with metric-specific valid and paired IDs.

For Gemini, panel and baseline Kendall $\tau_b$ are \newGeminiPanelTau{} and
\newGeminiBaselineTau{}; the paired baseline-minus-panel gap is \newGeminiGapTau{}
(CI [\newGeminiGapTauLow{}, \newGeminiGapTauHigh{}]) on \newGeminiPairedTauN{} valid pairs.
Expected top-1 is \newGeminiPanelTopOne{} versus \newGeminiBaselineTopOne{}, with paired gap
\newGeminiGapTopOne{} percentage points (CI
[\newGeminiGapTopOneLow{}, \newGeminiGapTopOneHigh{}]).

For GPT-4.1, panel and baseline Kendall $\tau_b$ are \newGptPanelTau{} and
\newGptBaselineTau{}; the paired gap is \newGptGapTau{}
(CI [\newGptGapTauLow{}, \newGptGapTauHigh{}]) on \newGptPairedTauN{} valid pairs.
Expected top-1 is \newGptPanelTopOne{} versus \newGptBaselineTopOne{}, with paired gap
\newGptGapTopOne{} percentage points (CI [\newGptGapTopOneLow{}, \newGptGapTopOneHigh{}]).
In Gemini, $\tau_b$ is undefined on \newGeminiBaselineRankUndefinedN{} baseline packages
and \newGeminiPanelRankUndefinedN{} panel packages. Marginal means therefore use different valid
sets; their difference is not the paired gap. The paired rank comparison uses their valid
intersection, while top-1 includes every complete package with expected credit for ties.
Baseline-minus-panel CIs exclude zero for the rank metrics, top-1 and captured lift in both
configurations; bottom-1 CIs include zero. These are pointwise conditional estimates.
Gamma and $\tau_b$ have distinct rows and denominators; neither is substituted for the other.
These conditional package-bootstrap intervals omit repeated-generation uncertainty and do not
compare model families causally. The GPT scalar-checkpoint analysis uses a separate collection protocol.
% AUTO-GENERATED by scripts/generate_recollection_assets.py
\begin{table*}[t]
\centering
\caption{Primary Gemini 3.1 Flash-Lite comparison on 188 complete packages from the 190-package selected cohort. Every retained variant has all planned observations in both arms. Counts are metric-specific valid panel/baseline/paired packages. Gap is baseline minus panel; decision gaps use percentage points. Percentile bootstrap CIs condition on this cohort and these recorded generations. Undefined coefficients are omitted with their denominators disclosed.}
\label{tab:new_gemini}
\small
\resizebox{\textwidth}{!}{\begin{tabular}{lcccc}
\toprule
Metric & Panel [95\% CI] & Baseline [95\% CI] & Paired gap [95\% CI] & $n_{valid}$ p/b/pair \\
\midrule
    Goodman--Kruskal $\gamma$ & 0.082 [-0.031, 0.192] & 0.381 [0.253, 0.501] & 0.268 [0.101, 0.434] & 185/137/136 \\
    Kendall $\tau_b$ & 0.082 [-0.030, 0.191] & 0.336 [0.214, 0.450] & 0.222 [0.061, 0.385] & 185/137/136 \\
    Spearman $\rho$ & 0.100 [-0.016, 0.214] & 0.371 [0.248, 0.488] & 0.236 [0.069, 0.405] & 185/137/136 \\
    Expected top-1 & 39.1\% [32.2, 46.3] & 54.8\% [49.4, 60.2] & 15.7 pp [7.1, 24.0] & 188/188/188 \\
    Expected bottom-1 & 43.4\% [36.4, 50.3] & 47.5\% [41.4, 53.6] & 4.2 pp [-4.1, 12.3] & 188/188/188 \\
    Captured lift & 0.458 [0.394, 0.522] & 0.591 [0.541, 0.641] & 0.133 [0.054, 0.210] & 188/188/188 \\
\bottomrule
\end{tabular}}
\end{table*}
\begin{table*}[t]
\centering
\caption{Primary GPT-4.1 comparison on 188 complete packages from the 190-package selected cohort. Every retained variant has all planned observations in both arms. Counts are metric-specific valid panel/baseline/paired packages. Gap is baseline minus panel; decision gaps use percentage points. Percentile bootstrap CIs condition on this cohort and these recorded generations. Undefined coefficients are omitted with their denominators disclosed. Azure requested three indexed choices per call; choices are not assumed independent.}
\label{tab:new_azure}
\small
\resizebox{\textwidth}{!}{\begin{tabular}{lcccc}
\toprule
Metric & Panel [95\% CI] & Baseline [95\% CI] & Paired gap [95\% CI] & $n_{valid}$ p/b/pair \\
\midrule
    Goodman--Kruskal $\gamma$ & 0.088 [-0.026, 0.200] & 0.345 [0.242, 0.445] & 0.257 [0.119, 0.391] & 188/188/188 \\
    Kendall $\tau_b$ & 0.088 [-0.026, 0.200] & 0.345 [0.242, 0.445] & 0.257 [0.119, 0.391] & 188/188/188 \\
    Spearman $\rho$ & 0.107 [-0.010, 0.222] & 0.373 [0.265, 0.475] & 0.266 [0.124, 0.404] & 188/188/188 \\
    Expected top-1 & 42.0\% [35.1, 48.9] & 61.7\% [54.8, 68.6] & 19.7 pp [10.6, 28.7] & 188/188/188 \\
    Expected bottom-1 & 43.1\% [36.2, 50.0] & 52.1\% [44.7, 59.0] & 9.0 pp [0.0, 17.6] & 188/188/188 \\
    Captured lift & 0.467 [0.400, 0.534] & 0.654 [0.590, 0.719] & 0.187 [0.103, 0.271] & 188/188/188 \\
\bottomrule
\end{tabular}}
\end{table*}

\paragraph{Post-hoc randomization-window sensitivity.}
Table~\ref{tab:window_sensitivity} excludes \windowGeminiExcludedN{} Gemini and
\windowGptExcludedN{} GPT-4.1 complete primary packages in the documented risk window,
leaving respectively \windowGeminiN{} and \windowGptN{} packages. Expected top-1 gaps remain
\windowGeminiGapTopOne{} percentage points (CI [\windowGeminiGapTopOneLow{},
\windowGeminiGapTopOneHigh{}]) and \windowGptGapTopOne{} points
(CI [\windowGptGapTopOneLow{}, \windowGptGapTopOneHigh{}]). Captured-lift gaps also remain positive
with pointwise CIs excluding zero.

Gemini's Kendall $\tau_b$ gap becomes \windowGeminiGapTau{}
(CI [\windowGeminiGapTauLow{}, \windowGeminiGapTauHigh{}]) on
\windowGeminiPairedTauN{} valid pairs; its tau-b and Spearman intervals include zero.
GPT-4.1's tau-b gap is \windowGptGapTau{} (CI [\windowGptGapTauLow{},
\windowGptGapTauHigh{}]) on \windowGptPairedTauN{} pairs. Thus the primary Gemini rank contrast
is not robust across this cohort restriction and all rank estimators. Gamma remains a distinct
conditional measure; it is not substituted for the uncertain tau-b result. Bottom-1 intervals
include zero for both models. These are post-hoc, pointwise intervals, not an equivalence test.
% AUTO-GENERATED by scripts/generate_window_sensitivity_assets.py
\begin{table*}[t]
\centering
\caption{Post-hoc sensitivity excluding the archive's documented randomization-risk window
(25 June 2013--10 January 2014 inclusive) from each model's complete primary cohort.
Primary reports and recorded generations remain unchanged. All six metrics are shown;
gaps are baseline minus panel, decision gaps are percentage points, and counts are metric-specific
valid pairs. Pointwise percentile bootstrap CIs condition on retained packages and generations;
remaining tests are not certified defect-free.}
\label{tab:window_sensitivity}
\small
\begin{tabular}{llcc}
\toprule
Model (complete packages) & Metric & Paired gap [95\% CI] & $n_{paired}$ \\
\midrule
Gemini (148) & Goodman--Kruskal $\gamma$ & 0.201 [0.010, 0.396] & 109 \\
Gemini (148) & Kendall $\tau_b$ & 0.157 [-0.030, 0.346] & 109 \\
Gemini (148) & Spearman $\rho$ & 0.170 [-0.022, 0.367] & 109 \\
Gemini (148) & Expected top-1 & 11.4 pp [1.7, 21.0] & 148 \\
Gemini (148) & Expected bottom-1 & 3.1 pp [-6.4, 12.6] & 148 \\
Gemini (148) & Captured lift & 0.091 [0.001, 0.181] & 148 \\
\midrule
GPT-4.1 (147) & Goodman--Kruskal $\gamma$ & 0.272 [0.118, 0.425] & 147 \\
GPT-4.1 (147) & Kendall $\tau_b$ & 0.272 [0.118, 0.425] & 147 \\
GPT-4.1 (147) & Spearman $\rho$ & 0.292 [0.133, 0.450] & 147 \\
GPT-4.1 (147) & Expected top-1 & 22.4 pp [12.2, 32.7] & 147 \\
GPT-4.1 (147) & Expected bottom-1 & 8.8 pp [-0.7, 18.4] & 147 \\
GPT-4.1 (147) & Captured lift & 0.213 [0.117, 0.307] & 147 \\
\bottomrule
\end{tabular}
\end{table*}

\begin{figure}[t]
\centering
\includegraphics[width=\linewidth]{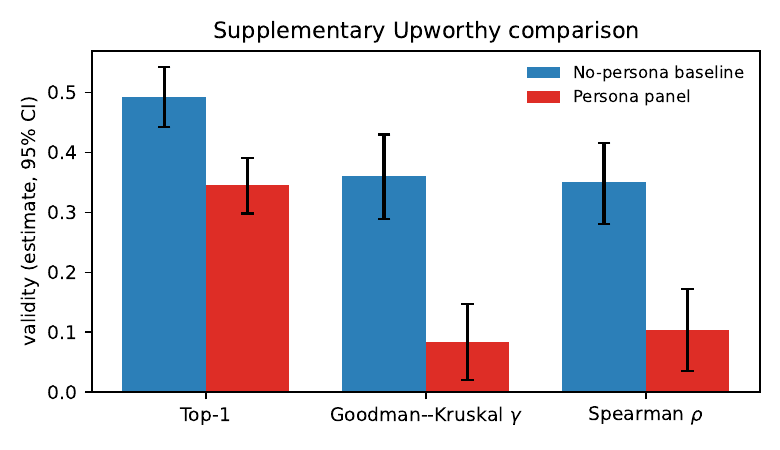}
\caption{Supplementary comparison under the top-versus-runner-up screen. Coefficient-specific valid counts are
unavailable. The displayed rank coefficient is gamma.}
\label{fig:comparison}
\end{figure}

\paragraph{Supplementary Upworthy comparison (RQ1).}
Table~\ref{tab:validity} and Figure~\ref{fig:comparison} give supplementary aggregate estimates.
The generic baseline has higher gamma (\baseGamma{} versus \goodmanGamma{}) and top-1
(\baseTopOne{} versus \topOneAcc{}). These are conditional descriptions under the top-versus-runner-up screen
and tie rule. Marginal CIs are not a paired test, and gamma is not assigned a calibrated
medium-effect interpretation.
% AUTO-GENERATED by scripts/generate_validity_table.py — DO NOT EDIT BY HAND.
\begin{table*}[t]
  \centering
  \caption{Supplementary Upworthy exploratory estimates under the top-versus-runner-up screen. Panel $n_{total}=399$; baseline $n_{total}=398$; correlation-specific valid counts are unavailable. The stored rank coefficient is Goodman--Kruskal gamma, not Kendall tau. Top-1 uses first-index tie breaking. Random top-1 reference 30.4\%. Marginal 95\% bootstrap CIs; the reported screen and decision tie rule are retained.}
  \label{tab:validity}
  \small
  \begin{tabular}{lccc}
    \toprule
    Method & Top-1 [95\% CI] & Goodman--Kruskal $\gamma$ [95\% CI] & Spearman $\rho$ [95\% CI] \\
    \midrule
    No-persona baseline & 49.2\% [44.2, 54.3] & 0.361 [0.289, 0.430] & 0.350 [0.281, 0.417] \\
    Persona panel & 34.6\% [29.8, 39.1] & 0.084 [0.020, 0.147] & 0.104 [0.035, 0.172] \\
    \bottomrule
  \end{tabular}
\end{table*}

\begin{figure}[t]
\centering
\includegraphics[width=\linewidth]{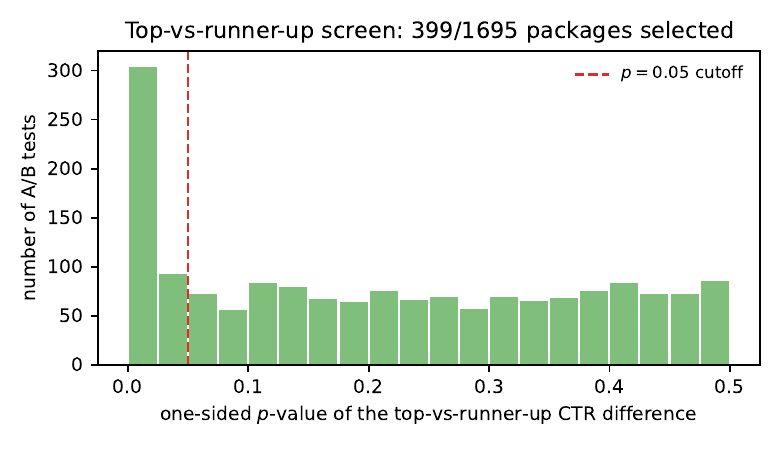}
\caption{Supplementary one-sided top-versus-runner-up screen selecting \legacySelectedN{} packages.
Selection of direction from observed rates invalidates a literal one-sided false-positive-rate interpretation.
Nonsignificant packages need not be unpredictable.}
\label{fig:reliability}
\end{figure}

Figure~\ref{fig:reliability} shows the supplementary screening statistic, not calibrated winner
probabilities. The supplementary all-package panel gamma is \sensGamma{} (CI
[\sensGammaCILow{}, \sensGammaCIHigh{}]), with the aggregate-report metric and parsing semantics.

\paragraph{Signal strength (RQ2).}
The pinned supplementary producer defines its separation index as $1-p$, where $p$ is the
one-sided observed-best-versus-runner-up value, and correlates it with the panel's first-index
top-1 hit. The stored Spearman association is \hThreeTopOne{} (CI
[\hThreeTopOneCILow{}, \hThreeTopOneCIHigh{}]) over \hThreeN{} packages, using $5{,}000$
bootstrap resamples. Its code requires every variant to have a score and within-package
Spearman to be defined, even for this top-1 analysis. The report alone cannot identify which
excluded packages lacked scores or had undefined ranks. This index mixes outcome separation
with exposure, and top-1 difficulty also varies with the number of variants; the association
is unadjusted for those factors. It is not evidence of an intrinsic predictability effect.
Missing supplementary predictions prevent recomputing the analysis under simultaneous screening
and expected tie credit; the primary selected-cohort experiment does not estimate this
all-package association.

\paragraph{Dataset comparisons (RQ3).}
Table~\ref{tab:replication} shows higher baseline gamma on the three Upworthy splits and a smaller
difference on MIND, where packages compare articles. On Reddit, the panel has higher top-1
(\redditPanelTop{} versus \redditBaseTop{}) and lift (\redditPanelLift{} versus \redditBaseLift{}).
The available summaries show different directions, but unreconciled collection and subset
differences prevent attributing this variation to the dataset alone. Reddit gamma
CIs include zero; this establishes neither equivalence nor absence of a title-related signal.
Posting time and visibility were not tested as explanations. Subset-selection and coverage
limitations further restrict interpretation. The aggregate summary does not record draw
counts or evaluated package IDs, so cross-dataset protocol equivalence is not independently
verified by those aggregates.
% AUTO-GENERATED by scripts/generate_replication_table.py — DO NOT EDIT BY HAND.
\begin{table*}[t]
  \centering
  \caption{Supplementary estimates under the top-versus-runner-up screen and first-index tie breaking. $n_{total}$ is panel/baseline coverage, not the correlation-specific valid count (unavailable). Lift is the fraction of the observed outcome-rate range captured by the selected variant (panel/baseline). The panel exceeds the baseline on Reddit lift and top-1 (discussed in text); near-zero gamma does not establish equivalence or unpredictability. Gap is the difference of reported marginal gamma means, not a paired contrast. The summary does not retain draw counts or evaluated package IDs; the exploratory row differs from Table~\ref{tab:validity}, and protocol equivalence cannot be verified from the aggregates alone.}
  \label{tab:replication}
  \small
  \begin{tabular}{lccccc}
    \toprule
    Dataset & $n_{total}$ p/b & Persona $\gamma$ & Baseline $\gamma$ & Gap & Lift p/b \\
    \midrule
    Upworthy exploratory (2013--15) & 399/398 & 0.079 & 0.361 & $+0.282$ & 0.45/0.58 \\
    Upworthy holdout (2013--15) & 402/401 & 0.100 & 0.398 & $+0.298$ & 0.48/0.56 \\
    Upworthy confirmatory (2013--15) & 1866/1864 & 0.070 & 0.393 & $+0.323$ & 0.44/0.55 \\
    MIND news (2019) & 74/74 & 0.177 & 0.214 & $+0.038$ & 0.47/0.51 \\
    Reddit titles (2008--12) & 600/599 & -0.009 & -0.026 & $-0.017$ & 0.45/0.33 \\
    \bottomrule
  \end{tabular}
\end{table*}

\paragraph{Seed, model, ties, and aggregation.}
The supplementary three-seed SDs (\stabGammaSD{} for gamma; \stabTopOneSD{} for top-1) use the
population convention over those seeds, not sample SDs or precise general-variability estimates.
The baseline has higher gamma on three tested Gemini tiers (Table~\ref{tab:cross_model_gap},
Appendix~\ref{sec:appendix}); evaluated IDs are not retained to verify a shared subset.
This cannot establish universal model or prompt robustness.

The tie report rounds scores before counting constant packages; the rank code uses exact
equality. That count cannot recover excluded gamma values. Undefined coefficients cannot be
silently assigned zero, and the reports do not justify a conservative-lower-bound claim for the gap.

Table~\ref{tab:h5agg} preserves six supplementary aggregation rules, all with lower reported gamma
than the baseline. Equal-persona and observation-weighted means can differ with missing draws.
The reports do not establish why their top-1 results differ, or whether bias, sampling noise,
scale, or another factor explains relative performance.
% AUTO-GENERATED by scripts/generate_h5_table.py — DO NOT EDIT BY HAND.
\begin{table}[t]
  \centering
  \caption{Supplementary aggregation comparison ($n_{total}=399$ for panel rules). The baseline has 398 covered packages; correlation-specific valid counts are unavailable. The mean row weights draw-collapsed personas equally, whereas the primary collector weights recorded observations equally. These estimands can differ when draws are missing; the stored reports do not identify the cause of their top-1 discrepancy. No tested rule reaches the baseline's reported gamma. This descriptive comparison does not identify a failure mechanism. Top-versus-runner-up screen, first-index tie breaking, 95\% marginal bootstrap CIs.}
  \label{tab:h5agg}
  \small
  \resizebox{\columnwidth}{!}{%
  \begin{tabular}{lcc}
    \toprule
    Aggregation & Goodman--Kruskal $\gamma$ [95\% CI] & Top-1 [95\% CI] \\
    \midrule
    No-persona baseline & 0.361 [0.289, 0.430] & 49.2\% [44.2, 54.3] \\
    \midrule
    Equal-persona mean & 0.084 [0.020, 0.147] & 34.8\% [30.1, 39.3] \\
    Panel median & 0.141 [0.076, 0.205] & 39.3\% [34.6, 44.1] \\
    Trimmed mean (20\%) & 0.081 [0.017, 0.144] & 34.3\% [29.6, 38.8] \\
    Per-persona $z$-mean & 0.064 [0.001, 0.128] & 33.6\% [29.1, 38.3] \\
    Borda (rank fusion) & 0.081 [0.018, 0.146] & 35.3\% [30.8, 40.1] \\
    Condorcet (pairwise) & 0.079 [0.014, 0.146] & 35.8\% [31.3, 40.6] \\
    \bottomrule
  \end{tabular}}
\end{table}

\paragraph{GPT-4.1 scalar-checkpoint analysis.}
The recovered checkpoint contains scalars, not original responses. Offline replay of the pinned
producer reproduces all published cross-family report fields within numerical tolerance.
Collection coverage is \xfN{} panel and \xfPairedN{} baseline/paired packages.
Thirty panel observations versus ten baseline draws and no generation seed make it
an unequal-budget comparison.

Table~\ref{tab:corrected_gpt} applies simultaneous screening and tie-aware estimators to those scores.
Coverage becomes \correctedPanelN{} panel, \correctedBaseN{} baseline, and \correctedPairedN{}
paired packages. Baseline Kendall $\tau_b$ is \correctedBaseTau{} versus \correctedPanelTau{}.
Paired CIs for the displayed gaps exclude zero, supporting a positive baseline gap in this
checkpoint analysis. It is not an independent replication or a budget-matched estimate, and does not validate
raw parsing. Gamma and tau-b remain separate. The supplementary Gemini observations are unavailable for applying these estimators; the
primary Gemini comparison uses its own retained generations and configuration.
% AUTO-GENERATED by scripts/generate_reanalysis_assets.py
\begin{table*}[t]
\centering
\caption{GPT-4.1 scalar-checkpoint analysis on recorded post-parser values. Simultaneous winner
screen and uniform expected tie credit; marginal percentile bootstrap CIs. Gap is baseline
minus panel on paired packages; top-1 gap is in percentage points. Counts are per-metric valid
panel/baseline/paired. Original responses are unavailable, so this does not validate the
collection parser. Sampling remains 30 panel observations versus 10 baseline draws per variant.}
\label{tab:corrected_gpt}
\small
\resizebox{\textwidth}{!}{\begin{tabular}{lcccc}
\toprule
Metric & Panel [95\% CI] & Baseline [95\% CI] & Paired gap [95\% CI] & $n_{valid}$ p/b/pair \\
\midrule
    Goodman--Kruskal $\gamma$ & 0.049 [-0.061, 0.159] & 0.325 [0.223, 0.429] & 0.270 [0.132, 0.407] & 190/189/189 \\
    Kendall $\tau_b$ & 0.049 [-0.061, 0.159] & 0.325 [0.222, 0.429] & 0.270 [0.132, 0.407] & 190/189/189 \\
    Expected top-1 & 39.5\% [32.6, 46.3] & 60.6\% [53.7, 67.7] & 20.9 pp [11.1, 30.4] & 190/189/189 \\
\bottomrule
\end{tabular}}
\end{table*}

\section{Discussion and Limitations}
The comparisons make a generic-reader baseline necessary for this task. A persona panel adds
assumptions about audience composition and behavior that require predictive validation.
The primary matched-score comparisons favor the baseline in valid rank comparisons, expected top-1
and captured lift, but do not establish a bottom-1 advantage. Gemini rank means exclude constant
outputs with explicit arm-specific counts; top-1 retains those packages through expected tie
credit. Supplementary Upworthy reports and the GPT checkpoint analysis also favor the baseline;
supplementary Reddit decision metrics favor the panel. Excluding the documented randomization-risk
period preserves the primary top-1 and lift contrasts but leaves Gemini tau-b and Spearman uncertain.
The ranking conclusion therefore depends on cohort and estimator, not just model configuration.

\paragraph{Mechanism remains open.}
The baseline may use general headline knowledge, while role-play may add unhelpful conditioning.
Neither explanation was isolated experimentally. Similar score spreads, selected ranking examples,
and rank-only aggregation failures cannot distinguish bias from noise or identify a mechanism.
The balanced, interleaved primary collection tests the panel/baseline comparison with retained raw
responses, but does not separately manipulate demographic content, role-play, or response noise.
It therefore does not by itself identify those mechanisms.
The archive measures clicks on Upworthy's website, while the prompt frames a social-media
reader in 2014 and omits the image. This platform/time framing is not an exact reconstruction
of the test audience. Intent versus engagement remains a construct-validity limitation~\citep{cronbach1955construct}.

\paragraph{Ground truth and generalization.}
The Upworthy archive documents a period of potentially invalid randomization
\citep{matias2024correction}; our selected cohort includes affected-window tests. Our predictions
rank recorded outcome rates, and neither an adjusted screen nor model responses establishes
causal headline effects. The post-hoc exclusion sensitivity addresses one known risk window,
not every possible allocation or image confound. Editorial stopping and shared content across
follow-up experiments~\citep{matiasUpworthy} add distinct risks: the screen does not correct
sequential monitoring, and the package bootstrap does not cluster related experiment IDs.
The preserved aggregate outcomes do not reconstruct the stopping trajectories, and dependence
has not been quantified by an article-level cluster analysis. Nominal intervals may understate
uncertainty. Simultaneous screening does not make outcome measurements certain. Exclusion changes the target population and relative task difficulty.
Failure to reject equality does not imply no predictor can recover an ordering.
MIND and Reddit are observational; correlated exposures, time, image, audience mix, and topic
may matter. Training contamination and generalization to current audiences or other formats
are not excluded by the archive corpus.

\paragraph{Reproduction limits.}
Aggregate reports allow reexport, not complete verification of their generating process.
Supplementary Gemini observations and some aggregate-report producers were not recovered. The GPT checkpoint
supports replay and partial reanalysis, but range-checking stored scalars cannot retroactively
validate the permissive parser. The primary strict-parser observations have complete raw lineage and matched within-model score
budgets; they are not mixed with supplementary scores. Azure multi-choice and Gemini single-choice
requests differ operationally. Returned aliases are not immutable model snapshots, and request
seeds do not guarantee determinism. A single collection session does not estimate provider drift
or generation variability. Content filtering can cause exclusions associated with content or arm;
the Azure technical pause also divides its collection in time. The outcome-selected cohort limits generalization.
Prompt robustness, deterministic demographic-to-panel construction, full Reddit subset lineage,
and preregistration chronology remain evidence gaps.

The artifact package distinguishes aggregate reexport, scalar-checkpoint analysis and raw-response
replay, so each reported comparison can be assessed within its available evidence.

\section{Conclusion}
This fixed demographic persona panel should be compared with a generic-reader LLM baseline.
The primary raw-retained comparison favors the baseline on valid rank comparisons, expected
top-1 and captured lift in both tested configurations. In the post-hoc sensitivity excluding the
archive's documented randomization-risk period, top-1 and lift gaps retain positive pointwise
intervals, but Gemini tau-b and Spearman intervals include zero. The rank advantage is therefore
not uniformly supported across these cohorts and estimators. In the primary comparison, expected top-1 baseline-minus-panel gaps
are \newGeminiGapTopOne{} percentage points for Gemini and \newGptGapTopOne{} for GPT-4.1.
Bottom-1 intervals include zero; a blanket advantage across metrics is not established.
Supplementary Reddit top-1 and lift favor the panel.

The comparison uses distinct rank metrics, expected tie credit, and explicit valid and paired counts.
We provide a simultaneous winner screen and an offline raw-replay artifact in the accompanying local
replication bundle, prepared for deposition. Rank estimates condition on valid coefficients, and all intervals condition on
selected packages and recorded generations. Related-test dependence and editorial stopping
remain unadjusted, limiting interval calibration and winner-screen interpretation. Supplementary Gemini observations remain unavailable. This study neither establishes
a general causal mechanism nor shows universal persona harm.

\section*{Data and Code Availability}
The package at \url{https://github.com/alemaiorano/sim-to-real-validity} includes sources,
exporters, supplementary aggregate reports, the fixed panel, and original corpus manifests.
Reexporting these reports reproduces their displayed numbers, not their generation from raw
observations. Original reports are hash-pinned and unchanged. GPT-4.1 checkpoint analyses,
input hashes, and offline replay verification are in \texttt{reanalysis/2026-10-31/}.
A separate replication bundle retains the GPT scalar checkpoint, exact
exploratory inputs, and the primary matched-budget experiment: all raw response bytes, request and
observation plans, immutable journal snapshots, coverage, package metrics, and source history.
Committed primary aggregate reports and hash provenance are in \texttt{recollection/2026-10-31/}.
Offline raw replay needs no provider account. Supplementary Gemini observations remain missing; primary
generations do not restore them. \texttt{REPLICATION.md} distinguishes these evidence tiers.
The accompanying replication bundle is prepared locally for deposition; it is not yet
available at the public repository URL.

\section*{Ethics and Conflict of Interest}
The study uses public archive data and synthetic personas; no new human subjects were recruited.
The author has a commercial interest in copy-simulation tools. This evaluates a study harness
rather than a specific product. Disclosure, adverse results, explicit limitations, and
available artifacts and the accompanying local bundle support scrutiny; an unverified preregistration chronology is not presented
as a safeguard.

\section*{AI Tools Disclosure}
This research used AI-assisted development tools for manuscript preparation and code development
under human oversight and accountability.
\begin{itemize}[leftmargin=*]
  \item \textbf{Language models:} Claude (Anthropic, via Claude Code) and GPT-family models
  (OpenAI, via Codex) supported code implementation, analysis, literature checks, and manuscript
  preparation and review. Automated manuscript feedback also supported the review. The Google
  Gemini and Azure GPT-4.1 configurations evaluated in the study are experimental components,
  as described in the methods and supplementary analyses; their outputs are study observations.

  \item \textbf{Web search:} MCP Tavily and web search tools supported literature review and
  fact-checking during manuscript preparation.
\end{itemize}

The author retains full responsibility for the scientific arguments, methodology, analysis,
citations, and conclusions. Numerical and literature claims are checked against retained
artifacts and primary sources; AI-generated suggestions and automated reviews do not constitute
independent empirical validation.

\bibliographystyle{plainnat}
\bibliography{references}

\appendix
\section{Persona Panel, Prompt, and Reproducibility Details}
\label{sec:appendix}
\paragraph{Fixed panel.}
The ten-persona JSON is the reproducible input. Demographic inspiration~\citep{holcomb2013newsuse}
is not a deterministic recipe for recreating behavioral descriptions. The unweighted panel is
not an estimated population sample.

\paragraph{First-person template.}
\begin{quote}\small
You are simulating a specific person reacting to a news headline on social media in 2014.\\
Person: \verb|{description}| (age \verb|{age_group}|, \verb|{gender}|, mainly on \verb|{platform}|,
news engagement: \verb|{news_engagement}|.)\\
Headline: \verb|"{headline}"|\\
As this exact person, how likely are you to click this headline? Respond ONLY with JSON:
\verb|{"click_intent":| \verb|<number between 0 and 1>}|.
\end{quote}
The primary panel condition uses this first-person template; the generic-reader condition uses
the baseline template below. Each is sent as a single user message, without a system message.
The exact templates, including spacing, quotation marks and implemented alternative framings,
are authoritative in the versioned collector. A prompt implementation alone does not
establish a completed experiment. The primary collection uses fixed 2014 framing for all
headlines, including those created outside that year; it is not a date-specific audience
reconstruction.

\paragraph{Generic-reader baseline template.}
\begin{quote}\small
You are a typical U.S. social-media news reader in 2014.\\
Headline: \verb|"{headline}"|\\
How likely are you to click this headline? Respond ONLY with JSON:
\verb|{"click_intent":| \verb|<number between 0 and 1>}|.
\end{quote}
Both templates retain the U.S./2014 framing. The baseline specifies a generic reader role;
its distinction from the panel is the absence of the fixed demographic profiles, not the
absence of all role conditioning. Literal spacing, persona fields, and request prompts are
retained in the collector and raw request plans. The interpolated persona fields are description,
age group, gender, platform, and news engagement, all shown in Table~\ref{tab:persona_panel}.
Political lean is stored metadata and is not separately interpolated.

\paragraph{Coverage and checkpoints.}
Coverage policies differ by analysis. The GPT scalar-checkpoint analysis requires at least
one recorded score for every variant, with means over available observations. The primary
matched-budget experiment requires all planned scores in both arms for a retained package. Terminal failures,
request/choice indices, and exclusions are preserved. Configuration and source hashes prevent
mixing protocols. Each Azure request has one seed shared by its three returned choices; there
is no invented per-choice seed. GPT scalar-checkpoint collection requests did not send a seed.

\paragraph{Raw artifact and replay.}
The primary collector appends and fsyncs a request record before dispatch and retains the exact
response bytes, status, usage, version, and parsing outcome afterward. Hash-chained journals
are synchronously mirrored. Unknown outcomes after interruption remain unknown and are not
silently regenerated. Offline analysis validates the request plan, input hashes, response hashes,
and journal chain, then reparses the same immutable snapshot whose hash identifies the report.
The replication bundle contains that snapshot, all inputs, request plans, aggregate reports,
source history, and the retained GPT scalar checkpoint. Replaying saved bytes
requires no provider account and does not assert that future generations will match.

\paragraph{Supplementary outcome construction.}
MIND aggregates the supplied news and behavior logs by article ID, counting appearances in
recommendation lists and click labels across users. Articles need at least 200 appearances;
within each subcategory, at most ten are retained by appearance count, with at least two per
package. Reddit aggregates identical titles within image/subreddit pairs, requires at least
300 total votes per title, and retains at most ten titles by vote count. Upvotes are the
numerator and total upvotes plus downvotes the denominator, regardless of the schema's
\texttt{clicks}/\texttt{impressions} names. These inputs cover the supplied logs, not a matched
exposure window or common audience. Original manifests and the pinned corpus builders specify
the transformations; missing evaluated IDs prevent fully reconciling the supplementary samples.

\paragraph{Supplementary pilot and model comparison.}
Tables~\ref{tab:model_comparison} and \ref{tab:cross_model_gap} retain the pilot and cross-tier
gap; shared package IDs are not retained in those summaries. Overlap of pilot CIs does not establish equivalence. The supplementary Upworthy comparison uses Gemini
3.1 Flash-Lite. No cost comparison is available from the pilot report.
% AUTO-GENERATED by scripts/generate_pilot_table.py — DO NOT EDIT BY HAND.
\begin{table*}[t]
  \centering
  \caption{Supplementary pilot with reported coverage $n_{total}=100$ packages per model. Evaluated IDs and coefficient-specific valid counts are unavailable; a shared subset cannot be verified. The intervals overlap; this is not an equivalence test. The stored pilot contains no cost measurements and does not establish cost-effectiveness. 95\% bootstrap CIs.}
  \label{tab:model_comparison}
  \small
  \begin{tabular}{lccc}
    \toprule
    Model & Top-1 & Goodman--Kruskal $\gamma$ & Spearman $\rho$ \\
    \midrule
    \texttt{gemini-2.5-flash} & 0.390 [0.29, 0.49] & 0.116 [-0.01, 0.24] & 0.133 [-0.01, 0.27] \\
    \texttt{gemini-3.1-flash-lite} & 0.330 [0.24, 0.42] & 0.035 [-0.10, 0.17] & 0.051 [-0.09, 0.19] \\
    \texttt{gemini-3.5-flash} & 0.390 [0.29, 0.49] & 0.122 [-0.01, 0.25] & 0.139 [-0.00, 0.28] \\
    \bottomrule
  \end{tabular}
\end{table*}

% AUTO-GENERATED by scripts/generate_crossmodel_table.py — DO NOT EDIT BY HAND.
\begin{table*}[t]
  \centering
  \caption{Supplementary cross-model comparison with reported coverage ($n_{total}=100$ per condition). The baseline has higher reported gamma on these three tested Gemini models. This does not establish a model-independent effect. Evaluated IDs and correlation-specific valid counts are unavailable; a common subset cannot be verified from this summary. Gap is the difference of reported marginal gamma means, not a paired contrast.}
  \label{tab:cross_model_gap}
  \small
  \begin{tabular}{lccc}
    \toprule
    Model & Persona $\gamma$ & Baseline $\gamma$ & Gap \\
    \midrule
    \texttt{gemini-3.1-flash-lite} & 0.035 & 0.401 & $+0.367$ \\
    \texttt{gemini-2.5-flash} & 0.116 & 0.219 & $+0.103$ \\
    \texttt{gemini-3.5-flash} & 0.122 & 0.404 & $+0.282$ \\
    \bottomrule
  \end{tabular}
\end{table*}

\paragraph{Headline-ranking examples and panel.}
Table~\ref{tab:error_examples} shows supplementary reported cases where the baseline found the observed winner
and the panel did not; they are not representative causal evidence.
Table~\ref{tab:persona_panel} presents the fixed panel descriptions. Full fields and grounding
notes remain in \texttt{data/persona\_panel.json}.
% AUTO-GENERATED by scripts/generate_supplementary_tables.py — DO NOT EDIT BY HAND.
\begin{table}[t]
  \centering
  \caption{Supplementary reported examples of observed winning headlines that the no-persona baseline ranked first but the persona panel did not (of \nBaseRightPersonaWrong{} such cases). These are selected illustrations, not evidence of a systematic causal mechanism.}
  \label{tab:error_examples}
  \small
  \begin{tabular}{p{5.1cm}cc}
    \toprule
    Real winning headline & CTR & \#var \\
    \midrule
    Bookmark This For When You’re Sad And Instantly Be 100 Times Happier & 1.38\% & 4 \\
    10 Amazing International Photos That Will Give You Some Perspective & 1.92\% & 4 \\
    If You Saw Someone Bullying A Kid For Being Gay Would You Speak Out... & 1.11\% & 2 \\
    If You Watch Hilary Swank Annoy A Whiny Senator, \$1 Will Go To Pun... & 1.67\% & 3 \\
    Seven Wonderful People Share Secrets That Are Killing Them.  This C... & 1.99\% & 4 \\
    A 9-Year-Old Asks McDonald’s CEO A Simple Question. She’s Not Lovin... & 3.23\% & 4 \\
    Did These Words Really Come Out Of Conan's Mouth? & 4.45\% & 3 \\
    We Use This Word So Often We Probably Don't Realize We're Saying It... & 1.68\% & 4 \\
    Did You Know You Can Go For A Swim In Google Earth? & 0.88\% & 3 \\
    Cookie Monster Is In Jail? I'm Actually Really Happy About The Reas... & 2.30\% & 4 \\
    \bottomrule
  \end{tabular}
\end{table}

% AUTO-GENERATED by scripts/generate_supplementary_tables.py — DO NOT EDIT BY HAND.
\begin{table*}[t]
  \centering
  \caption{Prompt fields of the fixed persona panel ($P{=}10$), inspired by Upworthy-era audience U.S. social-media demographics~\citep{holcomb2013newsuse}, with the verbatim behavioural descriptions used in the prompt. F/M denotes female/male. Political lean is retained as metadata in the replication package (\texttt{data/persona\_panel.json}) and is not separately interpolated.}
  \label{tab:persona_panel}
  \small
  \begin{tabular}{lp{2.3cm}p{1.6cm}p{8.9cm}}
    \toprule
    ID & Demographics & News engagement & Behavioural description (verbatim) \\
    \midrule
    p01 & 18-29, F, Facebook & incidental & 29-year-old woman, scrolls Facebook on her phone during breaks; clicks human-interest and feel-good stories shared by friends; rarely seeks news directly. \\
    p02 & 18-29, M, Twitter & active & 24-year-old man, heavy Twitter user, follows journalists and culture accounts; clicks surprising or contrarian takes; low tolerance for clickbait he's seen before. \\
    p03 & 30-49, F, Facebook & incidental & 38-year-old working mother, checks Facebook in the evening; drawn to headlines about family, health, and local impact; shares uplifting or outrage-inducing stories. \\
    p04 & 30-49, M, Reddit & active & 41-year-old man, browses Reddit and Facebook; skeptical of sensational headlines, clicks when a headline promises concrete information or a strong narrative payoff. \\
    p05 & 50-64, F, Facebook & incidental & 57-year-old woman, mostly on Facebook; clicks emotionally resonant stories and anything about social causes she follows; less drawn to internet-slang headlines. \\
    p06 & 50-64, M, Facebook & moderate & 60-year-old man, reads news on Facebook and cable; clicks political and economic headlines, distrusts overtly hyped 'you won't believe' framing. \\
    p07 & 18-29, F, YouTube & incidental & 22-year-old college student, video-first, on YouTube and Instagram; clicks visually evocative or identity-relevant headlines; very sensitive to authenticity. \\
    p08 & 30-49, M, Facebook & incidental & 45-year-old man, casual Facebook user; clicks curiosity-gap headlines that promise a quick interesting fact, especially maps, lists, and 'what happened next' stories. \\
    p09 & 65+, F, Facebook & moderate & 68-year-old retiree, active on Facebook with family; clicks heartwarming, nostalgic, or cause-driven stories; avoids headlines that feel like tricks. \\
    p10 & 30-49, F, Twitter & active & 34-year-old professional, follows news and advocacy on Twitter; clicks data-driven or justice-oriented angles; fatigued by repetitive viral formats. \\
    \bottomrule
  \end{tabular}
\end{table*}

\end{document}